\documentclass[letterpaper, 10 pt, conference]{ieeeconf}
\IEEEoverridecommandlockouts

\usepackage[backend=biber,
            hyperref=true,
            url=false,
            isbn=false,
            doi=false,
            backref=false,
            style=ieee,
            citestyle=numeric-comp,
            sorting=nyt,
            block=none]{biblatex}

\usepackage{flushend}
\usepackage{balance}
\usepackage{xcolor}
\usepackage{multirow}
\usepackage{float}
\usepackage{amsfonts}
\usepackage{siunitx}
\usepackage{textcomp}
\usepackage{graphicx}
\usepackage{xspace}
\usepackage{hyperref}
\usepackage{booktabs}
\usepackage{amsmath}

\usepackage{color,soul}

\newcommand{\algoName}{HAN\xspace}
\newcommand{\algoNameFull}{Hand-eye Action Networks\xspace}
\newcommand{\etal}{\xspace\emph{et al.}\xspace}

\renewcommand{\baselinestretch}{0.98}

\title{\LARGE \bf
Generalization Through Hand-Eye Coordination: An Action Space for Learning Spatially-Invariant Visuomotor Control
}

\author{
    Chen Wang$^{*}$, Rui Wang$^{*}$\thanks{Stanford Vision and Learning Lab, $^*$ denotes equal contributions}, Ajay Mandlekar, Li Fei-Fei, Silvio Savarese, Danfei Xu
}

\begin{document}

\maketitle
\thispagestyle{empty}
\pagestyle{empty}
\vspace{-20mm}

\begin{abstract}
Imitation Learning (IL) is an effective framework to learn visuomotor skills from offline demonstration data. However, IL methods often fail to generalize to new scene configurations not covered by training data. On the other hand, humans can manipulate objects in varying conditions. Key to such capability is hand-eye coordination, a cognitive ability that enables humans to adaptively direct their movements at task-relevant objects and be invariant to the objects’ absolute spatial location. In this work, we present a learnable action space, \algoNameFull (\algoName), that can approximate human’s hand-eye coordination behaviors by learning from human teleoperated demonstrations. Through a set of challenging multi-stage manipulation tasks, we show that a visuomotor policy equipped with \algoName is able to inherit the key spatial invariance property of hand-eye coordination and achieve zero-shot generalization to new scene configurations. Additional materials available at \url{https://sites.google.com/stanford.edu/han}

\end{abstract}

\section{Introduction}
Imitation Learning (IL) is a promising paradigm to acquire complex visualmotor skills by learning from a fixed set of expert demonstrations~\cite{Ijspeert2002MovementIW, Schaal1999IsIL, Billard2008RobotPB, Calinon2010LearningAR, englert2018learning,zhang2018deep}. However, IL
suffers from an important limitation - it is difficult for the
robot to generalize to new situations unseen in the training data, especially when learning to directly map image inputs to action outputs using neural networks~\cite{zhang2018deep}. This is because with finite training data and flexible function approximators, the robot may learn to focus on spurious correlations between the pixels and the demonstrated actions instead of the true intended behaviors. This limitation is especially pronounced for realistic multi-object manipulation tasks due to the large space of possible environment configurations. For example, a task of serving tea requires the robot to correctly manipulate a teapot and a mug starting from a \emph{combinatorial space} of the objects' initial configurations.  

A crucial ability that enables humans to manipulate objects under varying conditions is \emph{hand-eye coordination}, which is the cognitive ability of coordinating visual attentions and hand movements~\cite{johansson2001eye}. Such coordination coupled with the ability to switch visual attention at different stages of a task allows humans to adaptively direct their movements at task-relevant objects and be invariant to the objects' absolute spatial locations~\cite{johansson2001eye,bowman2009eye,rothkopf2007task}. For example, when serving tea, one would first locate the teapot and guide their hands toward the attended location. Once the teapot is grasped, the person would switch to attend to the mug and guide the teapot movements accordingly. 

\begin{figure}[t]
	\centering
    \includegraphics[width=1.0\linewidth]{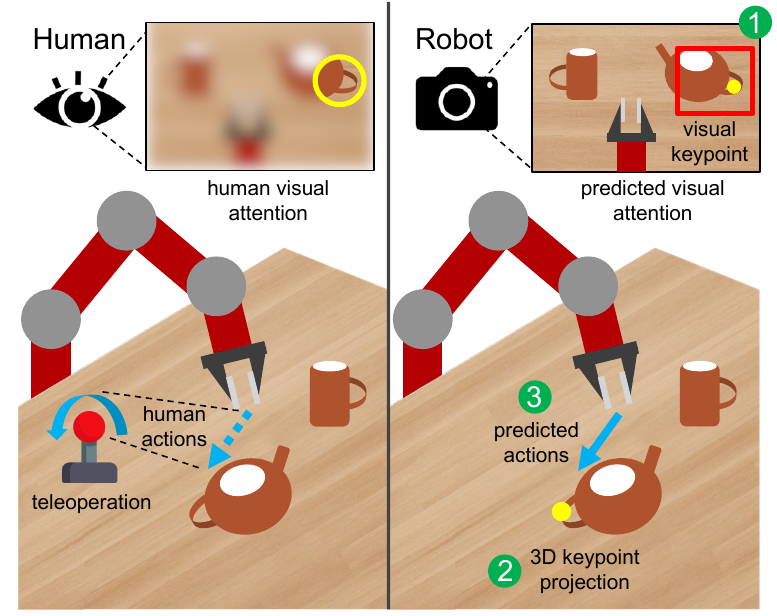}
	\caption{
	\textbf{(left)} Human exhibits hand-eye coordination when controlling the robot to complete a tea serving task: Human's gaze is fixated on the teapot handle when guiding the robot's action using a controller. Such coordinated movements is best explained by the relative locations of the gripper and the attended location rather than their absolute location, meaning that the movements is spatially-invariant.
	\textbf{(right)} Our method approximates  human's hand-eye coordination behavior through (1) generate a visual keypoint-based attention from the image observation (2) project the keypoint into the 3D scene and (3) guide the robot action by the attended target location.
	}
	\label{fig:pullfig}
\end{figure}

In the context of IL, humans that provide teleoperated task demonstrations also use hand-eye coordination to guide robot movements. In demonstrating the tea-serving task (illustrated in Fig.~\ref{fig:pullfig}), the operator's decision sequence is invariant to the initial workspace configuration: the demonstrated end effector movement can be explained by the relative location between the end effector and the teapot, rather than their absolute locations. Based on this observation, we hypothesize that we can attenuate the spurious correlations between the input images and demonstrated actions in visuomotor policies by enforcing human-like coordination of visual attention and end effector movements, enabling the policies to better generalize to new situations.
In this paper, we ask: can we enable imitation learning algorithms to recover the coordinated hand-eye movements of a human from a set of \emph{teleoperated demonstrations} of manipulation tasks? In other words, without knowing the true underlying cognitive process of the human operator, can we approximate the teloperator's hand-eye coordination and visual attention behaviors from a dataset of human-controlled robot actions and the corresponding image observations? 

To answer this question, we develop a \emph{learnable action space} called \algoNameFull (\algoName).
The goal of \algoName is to guide the robot's end effector movements directly using the 3D spatial locations selected by a learned visual attention model, and to learn such behavior directly from demonstration data. We illustrate the high-level idea in Fig.~\ref{fig:pullfig}. The key technical challenge in developing such an action space is that common visual attention models in computer vision research are based on 2D regions~\cite{mnih2014recurrent} or spatial maps~\cite{xu2015show}, whereas the robot end effector actions are in 3D. To bridge this gap, the primary component in \algoName is a \emph{3D visual attention network} that maps an input RGB image to a set of 3D keypoints relative to the robot's base coordinate frame. In addition, to mimic a human's ability to switch attention at different stages of a task, we introduce an \emph{attention switching network} to select a 3D target location from the candidate keypoints generated by the 3D attention layer. Finally, an \emph{action target network} learns to generate end effector action commands according to the selected 3D location.
Together, \algoName is a fully differentiable network within a deep imitation learning architecture, allowing end-to-end training with a simple behavior cloning objective, and is applicable to a wide range of manipulation tasks. 

Through a diverse set of robotic manipulation experiments including grasping, stacking and tool manipulation, we demonstrate the effectiveness of \algoName for learning visuomotor policies. More importantly, we show that \algoName inherits the key property of \emph{spatial-invariance} from human's coordinated hand-eye movements and is able to achieve strong zero-shot generalization to new environment initial configurations unseen in the training demonstrations.

The highlights of our work are as follows:
\begin{itemize}
    \item We develop a novel action space for learning human-like hand-eye coordination behaviors end-to-end from human-teleoperated demonstrations.
    \item To enable tight coupling between visual attention and robot actions, we develop a novel 3D attention mechanism that learns to generate 3D keypoints at task-relevant locations for guiding robot movements without direct keypoint supervisions.
    \item We evaluate on three simulated continuous control tasks of varying difficulties, and demonstrate that our action space enables a policy to generalize to tasks with unseen environment configurations in a zero-shot manner. We also show that the learned action space qualitatively exhibits human-like coordinated hand-eye movements.
\end{itemize}

\section{Related Works}

\textbf{Imitation learning.}
Imitation learning (IL) has been widely studied in the context of robotic manipulation tasks ~\cite{Ijspeert2002MovementIW, Schaal1999IsIL, Billard2008RobotPB, Calinon2010LearningAR, englert2018learning}. Typical IL formulations include inverse reinforcement learning (IRL) ~\cite{russell1998learning} which infers a cost function from expert demonstrations and behavioral cloning (BC) ~\cite{bain1995framework} which directly learns a policy mapping from observations to actions. This paper studies the problem of behavioral cloning. Recent works ~\cite{pomerleau1989alvinn, deep_imitation, Finn2017OneShotVI, mandlekar2020gti} use deep neural networks to map directly from image observations to action, and have proven effective for learning visualmotor skills for complex and diverse manipulation tasks. However, these policies tend to generalize poorly to new situations due to spurious connections between pixels and actions. This problem can be partially alleviated by learning from an online teacher~\cite{Ross2010ARO}, but this assumes that the expert is available on-demand to label actions, which can be infeasible for human supervisors~\cite{laskey2017comparing}, and the policy performance is still be limited by the coverage of the collected dataset.

\textbf{Visual attentions for policy learning.}
Drawing inspiration from human cognition~\cite{johansson2001eye,bowman2009eye,rothkopf2007task}, visual attention in computer vision enables a neural network model to focus on certain spatial locations of an image to extract information relevant for downstream tasks such as object recognition~\cite{mnih2014recurrent} and image captioning~\cite{xu2015show}. Recent works have attempted to equip interactive agents with attention~\cite{mott2019towards,attentionagent2020,zhangvisual,manchin2019reinforcement}. For example, Zhang\etal~\cite{zhangvisual} leverages human attention as a prior to extract relevant information from expert demonstration in imitation learning. While these works are effective in extracting task-relevant visual features, it is unclear how to leverage the extracted visual features for generating actions, especially when the 2D visual features and 3D robot actions reside in different spaces. Our paper introduces a new action space that directly uses the visually attended spatial locations to guide robot movements and can be trained end-to-end from human demonstrations. 

\textbf{Spatially-invariant action spaces.}
Our idea of spatially-invariant action space is inspired by a series of work by Zeng\etal and colleagues~\cite{zeng2018robotic,zeng2018learning,zeng2020tossingbot,zakka2020form2fit,song2020grasping,wu2020spatial}. Central to these works is the idea of grounding robot actions to a spatial action map, where each pixel in the action map indicates the Q-value of executing an action (e.g., grasping and pushing ~\cite{zeng2018learning}) at that spatial location. Because the action map is generated through a fully convolutional network, the convolution kernels can be shared across different locations on the map, enabling the robot to manipulate objects at locations outside of the training range. However, these methods are tied to top-down views with specially designed high-level actions such as top-down grasping or pushing, limiting their applicability to more fine-grained manipulation tasks. In contrast, our method can learn spatially-invariant policies with continuous end effector control, allowing our method to solve complex manipulation tasks such as tasks that require tool-use.

\textbf{Keypoint representations for control.}
Keypoints have been shown to be an effective representation for visuomotor policy learning~\cite{kulkarni2019unsupervised,florencemanuelli2018dense,manuelli2019kpam,manuelli2020keypoints,qin2020keto}. For example, kPAM~\cite{manuelli2019kpam} uses 3D keypoints as features for deep imitation learning. 
However, these works mainly consider keypoints as a feature space or a way to regularize deep network-based policies. We instead show that we can learn to ground 3D keypoints onto task-relevant locations in a scene and use the grounded keypoints to guide robot actions, enabling strong generalization to unseen observations. Our method uses no additional supervision other than the continuous control actions demonstrated through teleoperation.

\textbf{Reference frame selection for control.}
Learning to automatically select the reference frame for the robot action has also been used for reducing the dimensionality of the data and allowing additional generalization~\cite{muhlig2009automatic, sharma2020learning, franzeselearning}. The major difference between our method and these works is we are learning to localize the action attention directly from the broad vision space without the assumption of knowing the salient objects~\cite{muhlig2009automatic} or the ground truth positions~\cite{sharma2020learning}. In a sense, we believe that human demonstrations already contain rich spatial hints for reference frame selection, which has the potential to be learned without additional supervisions~\cite{franzeselearning}. Also, visual observation is a more generic information input for most of manipulation tasks.

\section{Problem Definition}
We consider a robot manipulation task as a Markov Decision Process (MDP), $\mathcal{M} = (\mathcal{S}, \mathcal{A}, \mathcal{T}, R, \gamma, \rho_0)$, with state space $\mathcal{S}$, action space $\mathcal{A}$, transition distribution $\mathcal{T}(s_{t+1} | s_t, a_t)$ , reward function $R(s_t, a_t, s_{t+1})$, discount factor $\gamma \in [0, 1)$, and initial state distribution $\rho_0$. At every step, the policy $\pi$ observes $s_t$, chooses an action $a_t = \pi(s_t)$, and observes the next state $s_{t+1} \sim \mathcal{T}(\cdot | s_t, a_t)$ and reward $r_t = R(s_t, a_t, s_{t+1})$. The goal is to learn an policy $\pi$ that maximizes the expected return $\mathbb{E}[\sum_{t=0}^{\infty} \gamma^t R(s_t, a_t, s_{t+1})]$.

In this work, we take a behavior cloning (BC) approach to the imitation learning problem. We assume access to a dataset of $N$ task demonstrations $\mathcal{D} = \{\tau_i\}_{i=1}^N$ where each demonstration is a trajectory $\tau_i = (s^i_0, a^i_0, s^i_1, a^i_1, ..., s^i_{T_i})$ that begins in a start state $s^i_0 \sim \rho_0(\cdot)$. The goal of BC is to train a policy $\pi_{\theta}(s)$ to clone the actions in the demonstrations with an objective $\arg\min_{\theta} \mathbb{E}_{(s, a) \sim \mathcal{D}} ||\pi_{\theta}(s) - a||^2$.

In this work, we aim to enable policies trained with the BC objective to generalize to states not in $\mathcal{D}$ without additional training data (zero-shot generalization). Specifically, we focus on generalizing to tasks with new initial distributions $\hat{\rho}_0$. As a simple example, we wish to enable a policy trained on lifting a cube from the center of the table to generalize to tasks in which the cube is initialized at the edge of the table.

We focus on learning visuomotor policies. Our policy takes RGB image observations as input and outputs end-effector actions in the form of relative position changes $\Delta(x, y, z)$ and a binary gripper open / close action. We defer full 6-DoF control to future works.

\begin{figure*}[ht]
	\centering
    \includegraphics[width=1.0\linewidth]{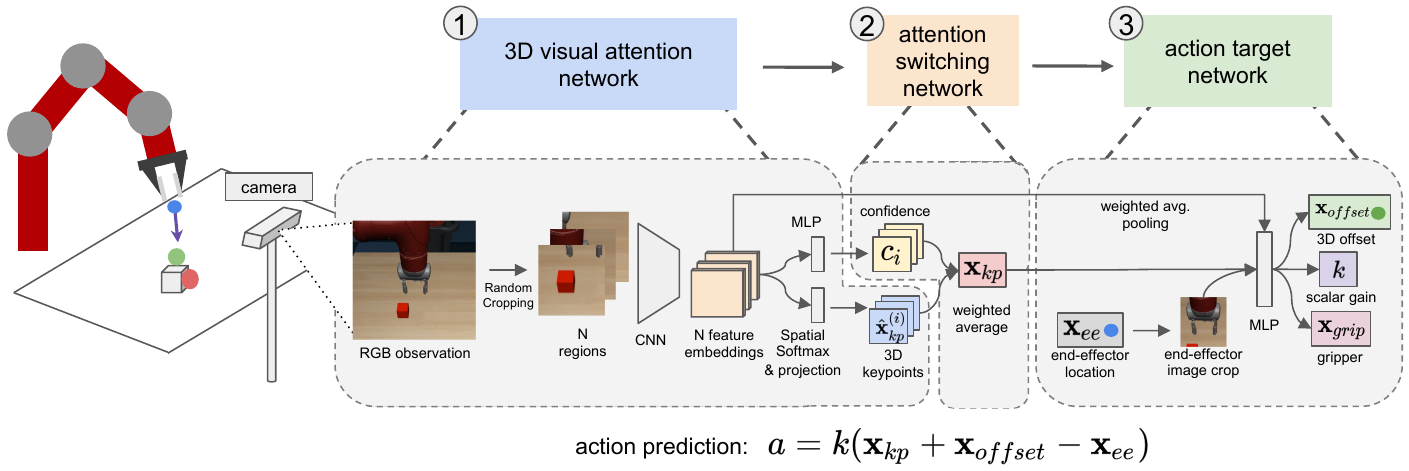}
	\caption{\textbf{Architecture Overview.} \algoName has three main components: (1) $N$ regions are randomly sample from the input image. Then a 3D visual attention network localizes a 3D keypoint $\hat{\mathbf{x}}_{kp}$ for each region. (Sec.~\ref{sec:3dattentionNet}) (2) An attention switching network generates a single target keypoint $\mathbf{x}_{kp}$ by aggregating the candidate keypoints through confidence-weighted sum. (Sec.~\ref{Sec:attention}) (3) The final local action target is set by moving the predict keypoint $\mathbf{x}_{kp}$ by a learned offset $\mathbf{x}_{offset}$. The network predicts offset $\mathbf{x}_{offset}$, control gain $k$, and the binary gripper open/close command $\mathbf{x}_{grip}$. The final output action $a$ is then calculated through the function on the bottom (Sec.~\ref{Sec:goalgrounding}).}
	\label{fig:model}
\end{figure*}

\section{Method}
\label{sec:method}
The primary technical contribution of this paper is a novel action space for imitation learning that enables trained visuomotor policies to generalize to new scene configurations without additional training data. The action space enables the agent to choose spatial locations in an image. These locations are projected to 3D and transformed into the robot frame, and are used to guide robot end effector actions. Through end-to-end training with an behavior cloning objective, the action space allows the agent to pay attention to objects of interest in the scene that best explain the actions taken by the demonstrator. In this way, this learnable action space tries to mimic the hand-eye coordination mechanism used by the human that provided demonstrations.

Our proposed action space consists of three core components: (1) a \emph{3D visual attention network} that focuses on regions relevant to the task and places \emph{3D keypoints} inside the regions, (2) an \emph{attention switching network} that attends to different keypoints depending on the current stage of the task, and (3) an \emph{action target network} that generates the robot end effector actions based on the attended 3D spatial locations. The three components are fully differentiable and can be trained end-to-end within a deep imitation learning architecture from a dataset of demonstrations. In the following, we motivate and describe each component in details. The overall architecture is illustrated in Fig.~\ref{fig:model}.

\subsection{3D Visual Attention Network}
\label{sec:3dattentionNet}
Common neural network-based visual attention mechanisms output 2D regions~\cite{mnih2014recurrent} or spatial maps~\cite{xu2015show}. However, since the robot's end effector actions are in the 3D space, it is difficult to recover the 3D location of the 2D attention in a differentiable pipeline.
To bridge this gap, we develop a 3D visual attention network that maps an input RGB image to a set of 3D keypoints relative to the robot's base coordinate frame. The network consists of an non-parametric 2D region proposal module and a learned 3D keypoint detector.

\textbf{2D region proposal for coarse-level attention:} Given an input image of size $H\times W \times 3$, we take $N$ fixed-size random crops of the image, each with size $H'\times W' \times 3$, where $H' < H, W' < W$. These regional crops can be considered as the candidates for \textit{coarse-level} attention for the subsequent attention switching network (Sec.~\ref{Sec:attention}) to select the region of interests (ROI) that are likely to contain task-relevant objects. While this is similar to the region proposal network (RPN) in the 2D detection framework FasterRCNN~\cite{ren2016faster}, there is no supervision on which regions contain objects - the proposals are completely random and serve as the basis for the subsequent attention switching mechanism.

\textbf{3D keypoint detection for refined attention:} To bridge the gap between the 2D regions and 3D robot movements, we learn a 3D keypoint detector to extract a 3D keypoint from each 2D region generated by the region proposal layer.
The keypoint detector is a deep convolutional network followed by a spatial softmax layer~\cite{finn2016deep}. 
The spatial softmax layer maps each region's convolutional feature map to a \textit{single} 2D keypoint $(u'_i, v'_i)$, with $u'_i \in [0, H'], v'_i \in [0, W']$ indicating the location of the keypoint relative to the $i$-th region. We convert each relative keypoint location to global pixel coordinates to obtain keypoints $\{(u_i, v_i)\}_{i=1}^N$ for all $N$ regions in the input image frame.

To convert these 2D keypoints to 3D, we use the depth values $\{d_i\}_{i=1}^N$ at each location, captured from a depth camera, and use the camera parameters and extrinsic transformation between the camera and the robot to transform keypoints in the camera frame $\{(u_i, v_i, d_i)\}_{i=1}^N$ to 3D keypoints in the robot frame $\mathbf{x}_{kp}=\{(x_i, y_i, z_i)\}_{i=1}^N$.

The 3D keypoints provide \textit{fine-grained} attention for localizing candidate task-relevant objects in the scene. For instance, when the robot is tasked to to grasp the teapot in Fig. \ref{fig:pullfig}, an ideal 3D keypoint should be grounded onto the handle of the teapot to guide the end effector movement. However, we still need to select a \emph{single} 3D location from the $N$ 3D keypoint candidates for guiding the robot action. Next, we describe how to make such a selection depending on the current stage of the task.

\subsection{Attention Switching Network}
\label{Sec:attention}

Everyday manipulation tasks are multi-stage, and each stage requires the robot to attend to different parts of the scene. In the tea serving example, after securing a stable grasp of the teapot, the robot needs to \emph{switch attention} from the teapot handle to the mug for the next stage of the task. We model such behavior with an attention switching network.

The main function of the attention switching network is to use the candidate keypoints from the 3D visual attention network (Sec.~\ref{sec:3dattentionNet}) to generate a single 3D location to guide the robot's next action. Instead of selecting exactly one keypoint, which would be non-differentiable, we take a weighted average of the candidates, where the weights are learned. 
The weights are computed by passing the convolutional feature maps for each region (Sec.~\ref{sec:3dattentionNet}) through a shallow multi-layer percetron (MLP) and softmax normalization.
The final output 3D location $\mathbf{x}_{kp}$ is simply a weighted sum of all the the candidate keypoints $\hat{\mathbf{x}}_{kp}$, i.e., $\mathbf{x}_{kp} = \sum_{i=1}^N c_i \hat{\mathbf{x}}_{kp}^{(i)}$, where $c_i$ is the normalized confidence of each candidate keypoint. 

\subsection{Action Target Network}
\label{Sec:goalgrounding}

Having described how to locate task-relevant objects in a scene using the 3D visual attention network and how to select a target 3D keypoint based on the current stage of the task, we now explain how to guide the robot action with the selected target keypoint.

The goal of the action target network is to map the predicted target keypoints to robot end effector actions that best match the demonstrated actions. While it is tempting to parameterize this mapping with a neural network, such flexible function approximators may prevent the model from learning meaningful coordination between the target keypoint and the action: empirically, we found that the network simply learns to ignore the target keypoint and use other input information instead (described later). 

On the other hand, the mapping should be expressive enough to fit the highly nonlinear relationship between the target and the end effector in complex manipulation tasks.
For example, to grasp an object top-down, the gripper needs to first move to a location above the object to avoid collision, and then grasp by approaching from above. In addition, human demonstrations often contain noises and suboptimal actions that cannot be explained by, e.g., a linear path.
To balance between over-parameterization and expressiveness, we propose to train a neural network to set constrained \emph{local action targets} relative to the target keypoint.
Specifically, the network sets action target location $\mathbf{x}_{target}$ relative to the keypoint $\mathbf{x}_{kp}$ by predicting a 3D \emph{spatial offset} vector $\mathbf{x}_{offset} \in \mathbb{R}^3$. The action target relative to the end effector position $\mathbf{x}_{ee}$ can thus be expressed as:
\begin{equation}
    \mathbf{x}_{target} = \mathbf{x}_{kp} + \mathbf{x}_{offset} - \mathbf{x}_{ee}
    \label{eq:offset}
\end{equation}

To generate the local end effector actions $a =\Delta(x, y, z)$, we use a learned scalar gain $k$ to modulate the action magnitude based on the target location relative to the end effector:
\begin{equation}
    a = k (\mathbf{x}_{kp} + \mathbf{x}_{offset} - \mathbf{x}_{ee})
    \label{eq:action_calc}
\end{equation}

$\mathbf{x}_{offset}$, $k$, and $\mathbf{x}_{grip}$ (gripper open / close) are predicted by an MLP. The inputs to this MLP are (1) keypoint position $\mathbf{x}_{kp}$ (2) end effector position $\mathbf{x}_{ee}$ (3) the visual features of a local image patch around the current gripper, and (4) the visual features of all the ROIs. These features provide both local and global information of the entire scene and allows the network to predict fine-grained control commands.

In practice, we constrain the range of the each coordinate in $\mathbf{x}_{offset}$ to be between -0.1 and 0.1. We also constrain the gain $k$ to be between 0 and 1. Empirically, we found these constraints to work well across the tasks we evaluated on. Such a parameterization both allows the network to fit the highly nonlinear demonstration trajectories and also prevent it from degenerating and ignoring the target keypoint. In the experiment section, we present both quantitative and qualitative evidence that the constrained action target allows \algoName to recover meaningful hand-eye coordination behaviors.

\subsection{Deep Imitation Learning}
The three components as described above constitutes our \algoNameFull. The networks are trained end-to-end with the behavior cloning objective given a set of demonstration data. The loss function has two components: (1) an L2 loss and (2) a Cosine similarity loss:

\begin{equation}
    L = ||a_t - a_t^{*}||^2 + \lambda \arccos(\frac{a_t^Ta^{*}_t}{||a_t||~||a^{*}_t||})
\end{equation}
where $a_t$ is the predicted four-dimensional action (first three from \eqref{eq:action_calc}, the last is the gripper action), the $a_t^{*}$ is the ground truth action from the demonstration. $\lambda$ is a constant balancing the two loss terms. The Cosine similarity loss improves directional alignment between the demonstrated and the predicted actions~\cite{zhang2018deep}.

\section{Experiments}
In this section, we seek to answer the following questions: (1) Does including \algoName in a deep imitation learning pipeline improve the task performance and zero-shot generalization ability? (2) Does our method approximate hand-eye coordination behavior? Specifically, does \algoName learn to focus on task-relevant regions and switch attentions at different stages of the task without direct supervision?
(3) How does each component of \algoName affect the final performance?

To answer these questions, we designed three table-top manipulation task domains with varying difficulties: from the most basic block lifting to the most complex multi-stage tool-using domain. We evaluate our method and all baselines on both training task distribution and zero-shot generalization to tasks with new initial configurations. We then analyze the role that each component of \algoName plays through a series of ablation studies. Finally, we provide qualitative analysis of the learned hand-eye coordination behavior.

\subsection{Task Setup}
All tasks are designed using MuJoCo~\cite{todorov2012mujoco} and the robosuite framework~\cite{robosuite2020}. The workspace consists of a Sawyer robotic arm in front of a table. The arm is controlled using an Operational Space Controller~\cite{khatib1987unified} in end effector positions. 
Environment observations are RGB images captured by a front-view camera rendered at $120\times 160$ resolution. We in addition make use of a depth camera to extract depth values for 3D keypoint generation (Sec.~\ref{sec:3dattentionNet}).

\begin{figure}[t]
	\centering
    \includegraphics[width=1.0\linewidth]{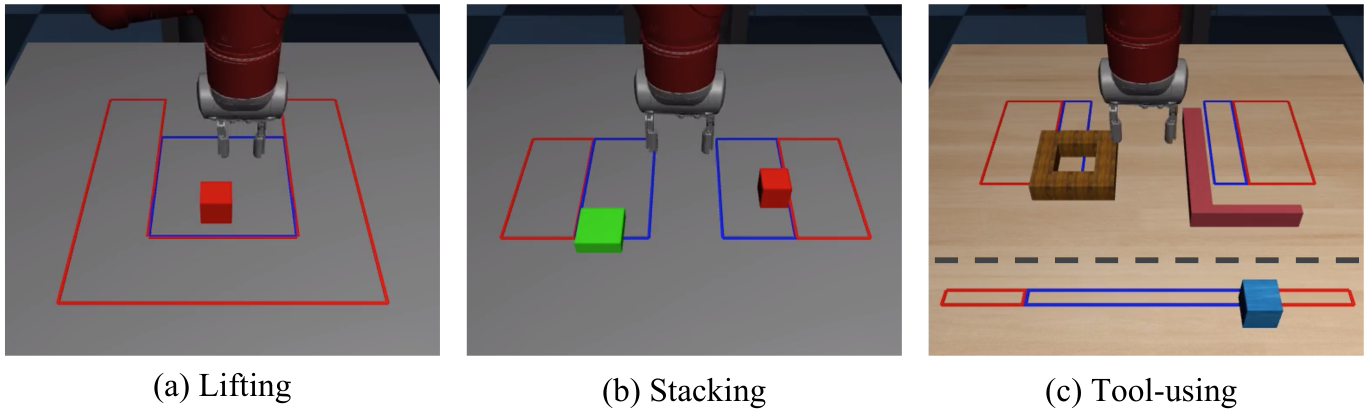}
	\caption{Initial states for the three manipulation tasks we evaluate on. During data collection, objects are initialized in the \emph{interpolation} regions, which are delineated by the red lines. We evaluate all methods with tasks that are initialized with both interpolation and \emph{extrapolation} regions, which are delineated by red lines. The dashed line in \emph{Tool-using} environment indicates the boundary beyond which objects are deemed unreachable by the robot, which requires the assistance of the tool.}
	
	\label{fig:exps}
	
\end{figure}

\textbf{Evaluation protocols:} To evaluate zero-shot generalization, we define \emph{interpolation} and \emph{extrapolation} regions for object initialization in each task, as shown in Fig.~\ref{fig:exps}.
We collect demonstrations only on tasks with objects initialized in the interpolation regions and report results on both interpolation and extrapolation tasks.

\begin{table}[]
\centering
\caption{\textbf{Quantitative Evaluation in the \textit{Lifting} environment}}
\setlength{\tabcolsep}{4mm}{
\begin{tabular}{l|llll}
\toprule
Demo type      & \multicolumn{2}{c}{50-Expert}                      & \multicolumn{2}{c}{200-Expert}                      \\ \hline
Eval region & \multicolumn{1}{c}{Int.} & \multicolumn{1}{c}{Ext.} & \multicolumn{1}{c}{Int.} & \multicolumn{1}{c}{Ext.} \\ \hline
BC-states      & 0.77                     & 0.2                      & 0.97                     & 0.5                      \\
BC-image\cite{florence2019self}       & 0.37                     & 0.1                      & 0.67                     & 0.13                     \\
HAN(mlp-ATN)    & 1.0                      & 0.47                     & 1.0                      & 0.57                     \\
HAN(no-ROI)    & 0.33                     & 0.07                         & 0.97                     & 0.1                     \\
HAN(no-con)    & 1.0                 & 0.53                         & 1.0                     & 0.67                     \\
HAN            & 1.0                      & \textbf{0.6}                      & 1.0                      & \textbf{0.73}                     \\ \bottomrule
\end{tabular}
}
\label{exp:lifting}
\end{table}

\begin{table}[]
\centering
\caption{\textbf{Quantitative Evaluation in the \textit{Stacking} environment}}
\setlength{\tabcolsep}{1.5mm}{
\begin{tabular}{l|llllllll}
\toprule
Demo type & \multicolumn{2}{c}{100-Expert}                     & \multicolumn{2}{c}{200-Expert}                      & \multicolumn{2}{c}{100-Human}                       & \multicolumn{2}{c}{200-Human}                       \\ \hline
Eval region & \multicolumn{1}{c}{Int.} & \multicolumn{1}{c}{Ext.} & \multicolumn{1}{c}{Int.} & \multicolumn{1}{c}{Ext.} & \multicolumn{1}{c}{Int.} & \multicolumn{1}{c}{Ext.} & \multicolumn{1}{c}{Int.} & \multicolumn{1}{c}{Ext.} \\ \hline
BC-states & 0.53                     & 0.1                      & \textbf{0.97}                     & 0.73                      & \textbf{0.67}                     & 0.07                     & 0.67                     & 0.13                     \\
BC-image\cite{florence2019self}  & 0.3                      & 0.03                     & 0.8                      & 0.13                      & 0.3                      & 0.03                     & 0.5                      & 0.1                      \\
HAN(mlp-ATN)  & 0.8                     & 0.47                     & 0.93                      & 0.7                       & 0.4                      & 0.13                     & 0.73                     & 0.23                     \\
HAN(no-ROI)  &  0.47                  &  0.13                   &  0.9                   & 0.13                & 0.27                     & 0.03                     & 0.5                      & 0.03                     \\
HAN(no-con)  & 0.93          & 0.53        &  0.9         & 0.63        & 0.53                & 0.17                & 0.43                 & 0.3                 \\
HAN   & 0.93                     & \textbf{0.63}                     & 0.93                     & \textbf{0.87}                      & 0.6                      & \textbf{0.33}                     & \textbf{0.83}                     & \textbf{0.57}                     \\ \bottomrule
\end{tabular}
}
\label{exp:stacking}
\vspace{-10pt}
\end{table}

\textbf{\emph{Lifting}:} As shown in Fig.~\ref{fig:exps}(a), the task is to grasp and lift a cube initialized at a random location by 10 cm. This task evaluates the generalization ability of all methods without introducing the challenge of multi-stage tasks.

\textbf{\emph{Stacking}:} The robot must pick up the red block and place it on top of the green plate, as shown in Fig.~\ref{fig:exps}(b). During training, the red block and the green plate are initialized in the interpolation region (blue). The corresponding extrapolation regions are adjacent to the interpolation regions (red). This task evaluates our method's ability to switch attention (red block then green plate) at different stages of the task.

\textbf{\emph{Tool-using}:} As shown in Fig.~\ref{fig:exps}(c), the robot must insert the blue cube into the wooden ring. However, the arm may not move past the dotted line, so the robot needs to  (1) grasp the tool, (2) use the tool to hook the blue block behind the dotted line, (3) grasp the block, and (4) insert the block into the wooden ring. Each object has its own interpolation and extrapolation region as shown in the figure. First, completing the task requires a variety of manipulation skills such as grasping, tool-using and insertion. Second, the task is long-horizon ($\sim$450 steps compared to $\sim$100 for \emph{Lifting} and $\sim 130$ for \emph{Stacking}). Finally, fetching the cube using the tool requires the policy to learn to manipulate one object with another instead of modeling only gripper-object interactions, testing the generality of our hand-eye coordination formulation.

\begin{figure*}[t]
	\centering
    \includegraphics[width=1.0\linewidth]{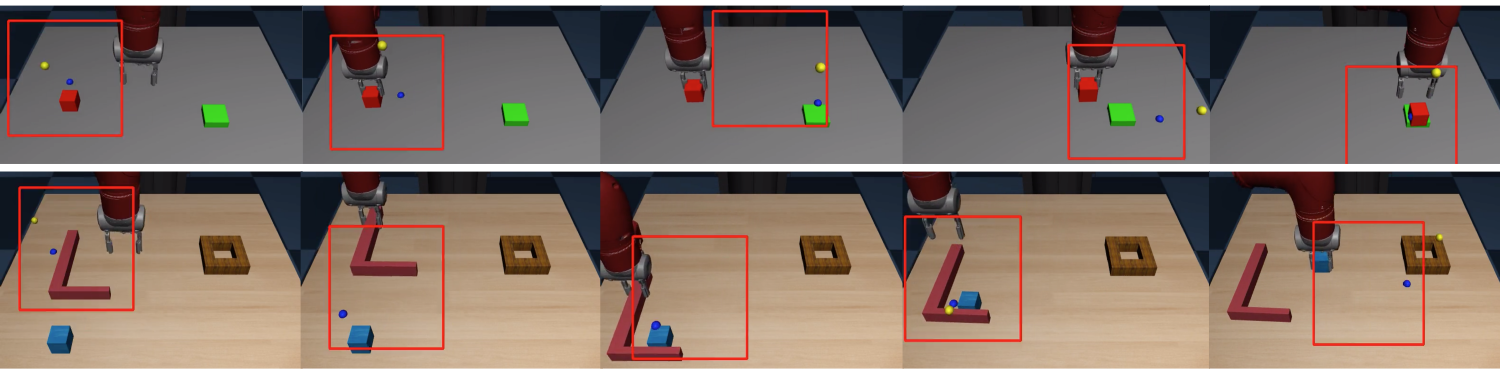}
	\caption{Key snapshots along the rollout of the \emph{Stacking} (top) and \emph{Tool-using} (bottom) domains, respectively. The blue spheres are the predicted keypoints. The yellow sphere indicates the location of the action target generated by the action target network (Sec.~\ref{Sec:goalgrounding}). Some of these action targets can be occluded by objects in the scene. The red boxes are the 2D bounding box with highest confidence scores (Sec.~\ref{Sec:attention}).}
	\label{fig:rollout}
	    \vspace{-5pt}
\end{figure*}

\subsection{Learning Setup and Baselines}
\label{sec:demo-baselines}
\textbf{Demonstrations:} We consider two sources of data: expert demonstrations and human demonstrations. Expert demonstrations are generated by a hard-coded policy with added Gaussian noise. The human demonstrations are recorded using RoboTurk~\cite{mandlekar2018roboturk, mandlekar2019roboturk}, which is a platform that allows users to demonstrate manipulation tasks through low-latency teleoperation with a smartphone interface. Learning from expert demonstrations examines whether we can recover the hand-eye coordination behaviors given ``clean data''. On the other hand, human demonstrations are often noisy and suboptimal. This means that not all demonstrated actions can be explained by hand-eye coordination. For example, we observed that a human teleoperator can take a curvy path to reach an object instead of a straight line. The sideways components of the actions cannot be explained by the attention. This challenging setup examines our method's ability to recover hand-eye coordination behaviors from noisy data.

\textbf{Baselines:} We compare our methods against five model variants to show the effectiveness of our design choices:
\begin{itemize}
    \item \textit{BC-states}: An MLP that takes ground-truth object poses and robot proprioception as input.
    \item \textit{BC-image}: A deep imitation learning model adapted from the end-to-end model variant of~\cite{florence2019self}. The baseline has the identical ConvNet backbone as our model, followed by a spatial softmax~\cite{ssmax} with 64 keypoints.
    \item \textit{\algoName (mlp-ATN)}: Our \algoName but with the action target network parameterized by an MLP that maps input features directly to actions instead of using our constrained action target formulation (Sec.~\ref{Sec:goalgrounding}).
    \item \textit{\algoName (no-ROI)}: Our \algoName but without the region-based attention. The 3D attention layer generates a single 3D keypoint directly from the entire input image instead of the ROIs. The 3D keypoints are then fed into our action target network to generate the final actions.
    \item \textit{\algoName (no-con)}: Our \algoName but without constraining the offset $\mathbf{x}_{offset}$ and scalar gain $k$.
    \item \textit{\algoName}: Our full model with all components in Sec.~\ref{sec:method}.
\end{itemize}

\subsection{Results}
We report performance of all methods as the maximum success rate achieved during training over 30 evaluation rollouts. In \emph{Lifting} (Tab.~\ref{exp:lifting}), with sufficient data (200 demos), all \algoName variants and BC-states are able to achieve near-perfect success rate in the interpolation region. In extrapolation, \algoName (mlp-ATN) outperforms the BC-image baseline significantly. Adopting our local action target formulation (Eq.~\eqref{eq:action_calc}) and imposing the constraints yields another absolute 10\% and 6\% gains, respectively. Note that the 2D region proposal is crucial for extrapolation (the no-ROI variant only achieves 0.1 success rate). This substantiates our main hypothesis that the spatial-invariance provided by the attention mechanism is crucial for zero-shot generalization. With less data (50 demos), \algoName achieves even greater gain over baselines and outperforms BC-states that has access to ground truth state in extrapolation with a wide margin (0.6 vs. 0.2). Similar trends are observed in \emph{Stacking} (Tab.~\ref{exp:stacking}) and \emph{Tool-using} (Tab.~\ref{exp:tool}).

\textbf{Hand-eye coordination} We further verify that our learned action space can approximate hand-eye coordination behaviors through qualitative visualization. In Fig.\ref{fig:rollout}, we visualize the predicted 3D keypoints, their corresponding 2D ROIs, and action targets in the key frames of two policy rollouts. For the top row (\emph{Stacking}), we highlight that immediately after the robot secured a grasp of the cube (frame 2), the 3D keypoint sharply switched to the green plate to proceed to the next stage of the task (frame 3). And the keypoint consistently locate the most relevant region to guide the robot actions. We observed similar behavior for \emph{Tool-using} (bottom row), where the network attends to the most relevant target object (tool, box, ring) at different stages of the task.

\begin{table}[]
\centering
\caption{\textbf{Quantitative Evaluation in the \textit{Tool Using} environment}}
\setlength{\tabcolsep}{4mm}{
\begin{tabular}{l|cccc}
\toprule
Demo type   & \multicolumn{2}{c}{100-Expert} & \multicolumn{2}{c}{200-Expert} \\ \hline
Eval region   & Int.           & Ext.          & Int.           & Ext.          \\ \hline
BC-states   & 0.13           & 0.03          & 0.23           & 0.0             \\
BC-image\cite{florence2019self}    & 0.3            & 0.03          & 0.33           & 0.07       \\
HAN(mlp-ATN) & 0.53           & 0.27          & 0.87           & 0.33          \\
HAN(no-ROI) & 0.0            & 0.0           & 0.0            & 0.0         \\
HAN         & \textbf{0.73}           & \textbf{0.43}          & \textbf{0.9}            & \textbf{0.6}           \\ \bottomrule
\end{tabular}
}
\label{exp:tool}
\vspace{-0pt}
\end{table}

\textbf{Human demonstrations} As noted in Sec.~\ref{sec:demo-baselines}, human demonstrations often contain noise and suboptimal actions. This requires our method to recover hand-eye coordination behaviors from noisy samples that may not conform to our assumptions. We evaluate learning with human demonstrations in \emph{Stacking} and show the results in Tab.~\ref{exp:stacking}. Compared to the other image-based methods, \algoName has the smallest performance drop between human and expert demonstrations and still outperforms all other baselines in both interpolation and extrapolation. Our observations for model ablations also still hold when learning from human data. This verifies that our approach is indeed capable of learning hand-eye coordination behavior from human data.

\section{Conclusion}
We presented \algoNameFull (\algoName), a fully differentiable policy network with a novel action space for learning robotic manipulation tasks through imitation learning. Through extensive experiments, we demonstrate \algoName's ability to solve challenging manipulation tasks and, more importantly, to generalize to unseen situations, a property not present in existing imitation learning algorithms. 

Although we mainly focused on imitation learning, other training frameworks such as reinforcement learning can potentially benefit from \algoName as well. We also plan to extend \algoName to handle rotational actions.

\section{Acknowledgement}

We acknowledge the support of Toyota Research Institute (``TRI''); this article solely reflects the opinions and the conclusions of its authors and not TRI or any other Toyota entity.
\begin{flushright}
\printbibliography
\end{flushright}
\end{document}